\pdfoutput=1
\documentclass[11pt]{article}

\usepackage[final]{acl}

\usepackage{times}
\usepackage{latexsym}
\usepackage{makecell}

\usepackage{url}

\usepackage[breaklinks]{hyperref}
\usepackage{tablefootnote}
\usepackage{amsmath} 
\usepackage{amssymb}
\usepackage{threeparttable}
\usepackage{graphicx}
\usepackage{pdflscape}
\usepackage{array}
\usepackage{expex}

\usepackage[T1]{fontenc}
\usepackage[utf8]{inputenc}

\usepackage{microtype}

\usepackage{inconsolata}

\usepackage{graphicx}

\title{Bias Attribution in Filipino Language Models: Extending a Bias Interpretability Metric for Application on Agglutinative Languages}

\author{
 \textbf{Lance Calvin Lim Gamboa\textsuperscript{1,2}},
 \textbf{Yue Feng\textsuperscript{1}},
 \textbf{Mark Lee\textsuperscript{1}},
\\
\\
 \textsuperscript{1}School of Computer Science, University of Birmingham,
\\
 \textsuperscript{2}Department of Information Systems and Computer Science, Ateneo de Manila University
\\
 \small{
   \textbf{Correspondence:} \href{mailto:email@domain}{llg302@student.bham.ac.uk}, \href{mailto:email@domain}{lancecalvingamboa@gmail.com}
 }
}

\begin{document}
\maketitle
\begin{abstract}
Emerging research on bias attribution and interpretability have revealed how tokens contribute to biased behavior in language models processing English texts. We build on this line of inquiry by adapting the information-theoretic bias attribution score metric for implementation on models handling agglutinative languages—particularly Filipino. We then demonstrate the effectiveness of our adapted method by using it on a purely Filipino model and on three multilingual models—one trained on languages worldwide and two on Southeast Asian data. Our results show that Filipino models are driven towards bias by words pertaining to \textit{people}, \textit{objects}, and \textit{relationships}—entity-based themes that stand in contrast to the action-heavy nature of bias-contributing themes in English (i.e., \textit{criminal}, \textit{sexual}, and \textit{prosocial} behaviors). These findings point to differences in how English and non-English models process inputs linked to sociodemographic groups and bias.
\end{abstract}

\section{Introduction}

As pretrained language models (PLMs) grow in scale and capability, research into the biased behaviors they exhibit continue to rise as well \citep{gallegos-etal-2024-bias,gupta-etal-2024-sociodemographic}. Improvements in their multilingual capacities, in particular, have been matched by studies investigating how fair multilingual and non-English models are (e.g., \citealp{steinnun-fridriksdottir-einarsson-2024-gendered}; \citealp{fort-etal-2024-stereotypical}; \citealp{ustun-etal-2024-aya}; \citealp{ibaraki-etal-2024-analyzing}). In these studies, NLP scholars from all over the globe take bias evaluation tools and methods initially developed for English and adapt them into multicultural contexts to detect how much bias multilingual PLMs demonstrate. These multilingual replications largely confirm the existence of safety and bias issues in models processing non-English texts. \citet{bergstrand-gamback-2024-detecting}, for example, found that the Norwegian models they experimented with prefer anti-queer statements over queer-friendly statements 68.27\% of the time on average. Meanwhile, \citet{huang-xiong-2024-cbbq} measured bias in Chinese question-answering models and discovered stereotypical associations between femininity, family duties, and career prejudices in some PLMs. 

Multilingual studies of bias, however, mostly focus on evaluation and, to a lesser extent, mitigation (e.g., \citealp{reusens-etal-2023-investigating}; \citealp{lee-etal-2023-square}) but do not engage the subjects of interpretability and explainability—that is, exploring the internal factors and mechanisms that influence biased decision-making among black-box PLMs \citep{liu2024devil}. Increasing the transparency of how these opaque models operate and improving our understanding of the roots of their biased behavior are important steps towards regulating their harmfulness and fostering public acceptance of these technologies \citep{xie-etal-2023-proto,lipton2016}. To these ends, \citet{gamboa-2024-interpretability} have developed an interpretability metric that explains how certain tokens contribute to bias in language models. Thus far, the method has only been applied on PLMs being evaluated on English bias tests and is yet to be extended to multilingual models handling non-English texts.

In this paper, we build upon their work by using the bias attribution score metric to analyze what tokens and semantic categories induce gender- and sexuality-biased tendencies within PLMs working on texts in Filipino, a language without high NLP resources \citep{joshi-etal-2020-state}. Examining gender- and sexuality-biased model behavior in Filipino holds value for three reasons. First is the swift adoption of AI technologies in Southeast Asia, where vulnerable minorities may be adversely affected by PLM biases and harms \citep{navarro2023generative,sarkar2023aiindustry}. Second is Filipino’s agglutinative morphology \citep{Gerona04032025,schachter2011tagalog}, which is distinct from English’s largely analytic morphology \citep{vangelderen2014} and therefore necessitates slight adjustments on tokenization-dependent methods such as bias attribution score calculation. The last reason pertains to idiosyncrasies in how gender, queerness, and related biases manifest in Filipino language and culture \citep{santiago2003balarila,cardozo2014comingout}, which may yield variations in how Filipino models manage gendered data as compared to English models. Indeed, our findings reveal that whereas the action-heavy topics of \textit{crime}, \textit{intimate relations}, and \textit{helping} prompt biased behaviors in models handling English \citep{gamboa-2024-interpretability}, PLMs processing Filipino can attribute their propensities for bias to words belonging to more concrete themes—e.g., those referring to tangible \textit{objects} and \textit{people}. 

 Our contributions are threefold:
 \begin{itemize}
    \item We are the first to leverage and adapt interpretability metrics in examining how individual tokens contribute to biased behavior in multilingual models working with non-English texts. 
    \item We adjust the derivation of the bias attribution score metric—initially used only for English—for use on agglutinative languages like Filipino.\footnote{Code available at \url{https://github.com/gamboalance/bias_attribution_filipino}}
    \item We uncover semantic categories that lead to biased decision-making in Filipino PLMs, thereby clarifying thematic areas in which these models should be used with caution and on which mitigation efforts should be focused. 
\end{itemize}

The remainder of this paper begins with a brief review of the literature regarding token-based attribution and interpretability in NLP (\ref{sec:related_work}). This review is followed by sections detailing our bias statement (\ref{sec:bias_statement}) and the methods we used—particularly, the dataset we chose, the models we inspected, and the attribution metric we used (\ref{sec:method}). The paper continues with the results of our analysis (\ref{sec:results}) and ends with our conclusions (\ref{sec:conclusion}). 

\section{Related Work}
\label{sec:related_work}
There are two categories of interpretability methods in machine learning: global and local explanation methods \citep{guidotti2019,lipton2016}. Global explanation methods shed light on the complete reasoning process employed by the model in arriving at all possible outcomes \citep{guidotti2019}. Global explanations seem to be rare for PLMs, especially among generative ones, because substantial variations in possible inputs and outputs make it hard to abstract a single interpretability tool, model, or heuristic that can generate explanations for all these possibilities. Instead, more common are local explanation methods, which examine data instances one by one and quantify how much a model’s prediction or output can be attributed to each individual input feature within a data point. Among language models, local explanations are achieved through the computation of token attribution scores. These scores indicate the degree of contribution each input token has on a PLM’s decision or generation \citep{attanasio-etal-2022-benchmarking,chen-etal-2020-generating-hierarchical}. 

Supervised learning models, such as those used to detect toxicity (e.g., \citealp{xiang-etal-2021-toxccin}), hate speech (e.g., \citealp{risch-etal-2020-offensive}), and misogyny (e.g., \citealp{attanasio-etal-2022-benchmarking,Godoy2021}), are the frequent subjects of token attribution studies that employ various mathematical techniques—e.g., linear approximations (e.g., \citealp{ribeiro2016}) and Shapley values (e.g., \citealp{chen-etal-2020-generating-hierarchical})—to calculate token attribution scores. \citet{gamboa-2024-interpretability} take inspiration from these classification-centric studies to propose a word-level attribution score that not only explains biased behavior but also works on language models trained under masked and causal learning paradigms. Specifically, their approach draws from an information-theoretic bias evaluation metric originally devised by \citet{steinborn2022information} and demonstrates that intermediary values needed to compute the bias metric can actually be used to explain word-level contributions to PLM bias as well. Their application of their proposed bias attribution score method on different PLMs allowed them to uncover how words linked to crime, intimacy, and helping can motivate models to behave with bias. Our study further refines their method for use on models handling agglutinative languages like Filipino and compares bias-inducing tokens among English and Filipino models. 

\begin{table*}
\small
  \centering
  \begin{tabular}{lll}
    \hline
    \textbf{Dimension} & \textbf{Sentence with More Bias} & \textbf{Sentence with Less Bias} \\
    \hline
    \makecell[tl]{Gender} & \makecell[l]{Laging pinagsasabihan ni Ginoong Reyes\\ang \textbf{babae} niyang katulong.\\ \textit{trans: Mr. Reyes always reprimands his} \\ \textit{\textbf{female} helper.}} & \makecell[l]{Laging pinagsasabihan ni Ginoong Reyes\\ang \textbf{lalaki} niyang katulong. \\ \textit{trans: Mr. Reyes always reprimands his} \\ \textit{\textbf{male} helper.}} \\ 
    \makecell[tl]{Sexual Orientation} & \makecell[l]{Tipikal siyang \textbf{bakla}, nakikipagtalik kung\\kani-kanino at naghahanap lagi ng bagong\\jojowain. \\ \textit{trans: He's a typical \textbf{gay} man, always for-} \\ \textit{nicating with everyone and always looking} \\ \textit{for a new boyfriend.}} & \makecell[l]{Tipikal siyang \textbf{lalaki}, nakikipagtalik kung\\kani-kanino at naghahanap lagi ng bagong\\jojowain. \\ \textit{trans: He's a typical \textbf{straight} man, always} \\ \textit{fornicating with everyone and always look-} \\ \textit{ing for a new boyfriend.}} \\
    \hline
  \end{tabular}
  \caption{\label{table:sample_sentences}
   Challenge prompt pairs in Filipino CrowS-Pairs. Distinguishing or modified tokens in each statement are in bold for emphasis. Linguistic glosses for the prompts are in Appendix \ref{app:glosses}.
  }
\end{table*}

\begin{table*}
\small
  \centering
  \begin{threeparttable}
  \begin{tabular}{lccccc}
    \hline
    \textbf{Model} & \thead{\textbf{Training}\\\textbf{Paradigm}} & \textbf{Language} & \thead{\textbf{Gender}\\\textbf{Bias Score}} & \thead{\textbf{Sexuality}\\\textbf{Bias Score}} & \thead{\textbf{Overall CP}\\\textbf{Bias Score}}\\
    \hline
    \texttt{gpt2} & causal & languages worldwide & 53.43 & 68.49 & 58.82 \\
    \texttt{roberta-tagalog-base} & masked & Filipino & 53.43 & 73.97 & 60.78 \\
    \texttt{sea-lion-3b}\tnote{a} & causal  & English \& Southeast Asian languages & 74.81 & 67.12 & 72.06\\
    \texttt{SeaLLMs-v3-7B-Chat}\tnote{b} & causal  & English \& Southeast Asian languages & 51.14 & 52.06 & 51.47 \\ 
    \hline
  \end{tabular}
  \caption{Models examined, their properties, and their bias scores as evaluated vis-a-vis Filipino CrowS-Pairs (CP). An unbiased model would have a score of $50.00$.}
  \begin{tablenotes}
  \item[a] SEALION: Southeast Asian Languages In One Network.
  \item[b] SEALLMs: Southeast Asian Large Language Models
  \end{tablenotes}
  \label{tab:models}
  \end{threeparttable}
\end{table*}

\section{Bias Statement}
\label{sec:bias_statement}
Conceptually, we ascribe to the notion of PLM bias as disparities in model performance associated with or arising from input data containing different sociodemographic attributes \citep{gallegos-etal-2024-bias,gupta-etal-2024-sociodemographic}. Operationally, we define bias as a violation of the \textit{equal social group associations} fairness condition specified by \citet{gallegos-etal-2024-bias}. A model fulfills \textit{equal social group associations} if non-demographically related words are equally likely to be chosen or generated in contexts relating to distinct social groups. For example, in a fair model, the word teacher would have an equal probability of being generated for the stems \textit{The boy grew up to be a…} and \textit{The girl grew up to be a…} Consequently, our operationalization of bias deems as unfair models which systematically prefer to associate certain neutral concepts with particular social groups. Concretely, we quantify this using Filipino CrowS-Pairs and bias metrics derived from comparing token probabilities—all of which we discuss with more detail in the next section.

This conceptualization and operationalization of PLM bias enables our study to elucidate the representational harms of models handling Filipino texts. Representational harms result from models perpetuating stereotypes about marginalized groups through generating unfavorable depictions about them or associating them with negative traits \citep{blodgett-etal-2020-language,crawford2017trouble}. Models that consistently link neutral but stereotypical concepts with certain demographics are culpable of committing such harms. Our analysis focuses on the potentially detrimental impacts of biased language model deployment on historically disadvantaged gender and sexuality groups in the Philippines—e.g., the \textit{babae} (the female), the \textit{bakla} (the non-heterosexual man), and the \textit{tomboy} (the non-heterosexual woman) \citep{velasco2022tomboy,garcia1996phgay,santiago1996}. 

\section{Method}
\label{sec:method}
\subsection{Data}
Bias evaluation benchmarks facilitate the measurement, examination, and comparison of biased behavior across language models. They are also a prerequisite to an interpretable analysis of model bias through the bias attribution score metric \citep{gamboa-2024-interpretability}. We use the Filipino CrowS-Pairs dataset to probe bias and explore bias interpretability among multilingual PLMs handling Filipino. Adapting the English CrowS-Pairs \citep{nangia2020crows} benchmarks to the Philippine setting, Filipino CrowS-Pairs is composed of 204 challenge prompt pairs that assess for two bias dimensions: gender and sexual orientation \citep{gamboa-lee-2025-filipino}. Each pair is made up of two minimally different statements: one conveying a stereotype or bias, and another expressing a less biased sentiment. As shown in Table \ref{table:sample_sentences}, these sentence pairs vary by only one or a few social attribute words, which modify the meaning and degree of bias of a statement when altered. 

Models that repeatedly judge biased statements as more linguistically probable over less biased counterparts are presumed by these benchmarks to hold stereotypes and prejudices learned from pretraining data. Given that Filipino CrowS-Pairs was developed with careful consideration of peculiarities in Philippine language and culture, it may be assumed that its resulting bias evaluations and metrics are contextually and culturally appropriate and relevant.

\subsection{Models}
We analyze bias interpretability across four models capable of processing Filipino. We examine both masked and autoregressive Transformer-based models, which are currently demonstrating the best performances in multilingual benchmarks \citep{zhao2024llamaenglishempiricalstudy,huang-etal-2023-languages}. We also look into models with different language compositions in their pretraining data: \texttt{roberta-tagalog-base} was trained on purely Filipino data \citep{cruz-cheng-2022-improving}, \texttt{sea-lion-3b} and \texttt{SeaLLMs-v3-7B-Chat} were trained on data in English and Southeast Asian languages \citep{aisingapore2023sealion,zhang2024seallm3}, and \texttt{gpt2} was trained on languages worldwide \citep{radford2019language}. Among these models, the \texttt{sea-lion-3b} model was found to be the most biased when tested against the entire Filipino CrowS-Pairs benchmark, while \texttt{roberta-tagalog-base} was found to be the most homophobic as evaluated using only the \textit{sexual orientation} subset of Filipino CrowS-Pairs. Table \ref{tab:models} provides a summary of the models we analyzed, their properties, and their bias as measured using Filipino CrowS-Pairs.

\subsection{Bias Attribution}
To examine how individual tokens contribute to biased model behavior, we use the bias attribution score proposed by \citet{gamboa-2024-interpretability}. This interpretable metric is computed using the equation below.

\begin{equation}
\resizebox{0.45\textwidth}{!}{$
b(u) = \sqrt{\text{JSD}(P_{u,\text{more}} \parallel G_u)} - \sqrt{\text{JSD}(P_{u,\text{less}} \parallel G_u)}
$}
\end{equation}

In this equation, the bias attribution score is denoted by $b(u)$ or the \textbf{b}ias of each \textbf{u}nmodified token in a CrowS-Pairs challenge pair. Unmodified tokens are the words shared by both sentences in a pair—e.g., \textit{tipikal} (\textit{typical}), \textit{nakikipagtalik} (\textit{fornicating}), and \textit{bagong} (\textit{new}) in the second example in Table \ref{table:sample_sentences}—and are distinguished from modified tokens, or the attribute words by which the sentences differ—e.g., \textit{lalaki} (\textit{straight man}) and \textit{bakla} (\textit{gay man}) in the same example. At a conceptual level, $b(u)$ calculates token-level bias contribution by comparing the probability of an unmodified token appearing in a stereotypical context (i.e., the biased statements in Table \ref{table:sample_sentences}) and the probability of the same token appearing in a less stereotypical context (i.e., the less biased statements). In the CrowS-Pairs bias evaluation paradigm, tokens that are more likely to appear in the stereotypical context directly contribute to a PLM preferring a biased sentence over a less biased one and increasing the model’s overall bias score.  

At a mathematical and pragmatic level, the bias attribution score method compares token probabilities in biased and less biased contexts by first obtaining $P_{u,more}$ and $P_{u,less}$. $u$ is the unmodified token whose bias attribution score is being calculated, and $P_{u,more}$ is the probability distribution computed by the model for \texttt{<MASK>} when the token is masked within the \textit{more} stereotypical context. For example, if we were determining the bias attribution score of \textit{fornicating} in the English translation of Table \ref{table:sample_sentences}’s second example, $P_{fornicating,more}$ would correspond to the distribution of probabilities the model assigns to each word in its vocabulary with respect to their likelihoods of filling \texttt{<MASK>} in the prompt \textit{He’s a typical gay man, always} \texttt{<MASK>} \textit{with everyone and always looking for a new boyfriend.}

Conversely, $P_{u,less}$ is the probability distribution provided by the model for \texttt{<MASK>} when the unmodified token is masked in the \textit{less} stereotypical context. Continuing the example above, $P_{fornicating,less}$ is the distribution enumerating the probabilities of each word in the model vocabulary filling \texttt{<MASK>} in \textit{He’s a typical straight man, always} \texttt{<MASK>} \textit{with everyone and always looking for a new boyfriend.}

Given that the distributions were conditioned on dissimilar bias contexts, it is expected that they will each assign different probability values to the model’s vocabulary—including the word whose bias attribution score is being calculated. For example, $P_{u,more}$ might assign \textit{sleeping} a probability of $0.89$ while $P_{u,less}$ might assign it a probability of $0.75$ because the model associates \textit{fornicating} more strongly with the word \textit{gay} (which is found in the more stereotypical context) than with the word \textit{straight} (found in the less stereotypical context).

With these differences in probabilities, one distribution also becomes naturally closer to the ground truth compared to the other distribution. In the example above, $P_{u,more}$ is closer to the truth because it assigns a higher probability ($0.9$) to the correct and relevant token (\textit{fornicating}). This indicates that \textit{fornicating} is more likely to be generated by the model in the \textit{more} biased condition than the \textit{less} biased condition. As such, \textit{fornicating} also makes it more likely for the model to generate or choose the more biased statement than the less biased statement. 

The next step in quantifying this contribution is to measure and compare the distances of the two distributions with the ground truth $G_u$, given by a one-hot distribution in which the probability of the relevant token $u$ is $1$ and the probability of every other token in the model vocabulary is $0$. The distances are computed by the Jensen-Shannon distance (JSD) formula from information theory \citep{lin1991jsd,endres2003jsd} and are subtracted from each other.

A resulting bias attribution score of less than $0$ indicates that the distance between the probability distribution under the more stereotypical context ($P_{u,more}$) is smaller and closer to the ground truth than the distance between $P_{u,less}$ and $G_u$. A negative bias attribution score may thus be interpreted as signaling that the relevant token is more probable in a biased context and consequently induces the PLM to select or generate more stereotypical statements. Conversely, a positive bias attribution score would signal the opposite: that the token pushes a model to act with less bias and prefer less stereotypical utterances. While the bias attribution score’s sign signifies a token’s direction of influence towards model bias, its magnitude represents the strength of this influence. 

\subsection{Bias Attribution for Agglutinative Languages}
For a dominantly analytic language like English, the bias score attribution method described above can be implemented in a straightforward manner. In analytic languages, an individual word often carries just one or a few concepts, rarely uses affixes, and is therefore relatively shorter in nature compared to words in synthetic and agglutinative languages \citep{Payne_2017}. This morphological typology of the English language allows PLM tokenizers to treat most English words as individual tokens. As such, in applying the bias attribution score method on English, each token’s $b(u)$ score often corresponds to a unique word’s score as well. 

The interpretability approach, however, becomes more complicated for agglutinative languages like Filipino, where a singular word can contain multiple affixes and concepts and are therefore longer in nature \citep{Payne_2017}. \textit{Fornicating}, for example, translates to \textit{nakikipagtalik} in Filipino. \textit{Nakikipagtalik} can be broken down or tokenized into five morphemes: \textit{na-}, \textit{ki-}, \textit{ki-}, \textit{pag-}, and \textit{-talik}, in which \textit{talik} is the root meaning \textit{intimate}, \textit{pag-} is a prefix indicating an \textit{action}, and nakiki- are a combination of prefixes denoting the present progressive and the performance of an action with another entity. Roughly corresponding to \textit{currently being intimate with someone}, \textit{nakikipagtalik} can therefore receive five different $b(u)$ scores for each of its subcomponent morphemes when subjected to a PLM tokenizer and the bias attribution score method described in the previous section. To resolve this complexity, we implement an additional step to the method proposed by \citep{gamboa-2024-interpretability}: for words which are further divided into tokens by the model tokenizer, the bias attribution score is given by the mean of the scores of its component subwords, which corresponds to the following:
\[
b(u) = \frac{1}{n} \sum_{i=1}^{n} b(t_i)
\]

where:
\begin{itemize}
    \item \textit{u} is the complete word whose attribution score is being calculated,
    \item $t_1, t_2, \ldots, t_n$ are the tokens resulting from tokenizing $u$, and
    \item $b(t_i)$ is the bias attribution score function applied to token $t_i$.
\end{itemize}

\subsection{Semantic Analysis}
To examine the semantic categories of words inducing biased behavior in Filipino PLMs, component words of Filipino CrowS-Pairs were first translated to English using the googletrans package and then semantically tagged using the pymusas package. pymusas is a semantic tagger that can characterize the semantic fields a word belongs to \citep{rayson2004ucrel}. Similar to \citet{gamboa-2024-interpretability}, we remove from our analysis words that comprise less than 1\% of the dataset’s total word count (i.e., words that occur less than $n=10$ times). In the next section, we report the semantic categories with the most bias-contributing tokens in terms of proportion.

\section{Results and Discussion}
\label{sec:results}

\subsection{Bias Attribution in Filipino}
Tables \ref{tab:bias_attrib1} and \ref{tab:bias_attrib2} show how the adjusted bias attribution score method is useful in providing interpretable explanations for the biased behavior of models handling Filipino. Table \ref{tab:bias_attrib1}, in particular, outlines how the shared tokens in Table \ref{table:sample_sentences}’s first example contributed to RoBERTa-Tagalog opting for the more biased statement over the less biased alternative. Among these tokens, the words \textit{pinagsasabihan} (\textit{reprimand}), \textit{laging} (\textit{frequently}), and \textit{katulong} (\textit{helper}) had negative bias attribution scores, suggesting that these contributed to the model’s biased behavior in this context. It is possible that the combination of these tokens motivated the model to decide that it is more probable for the statement to be referring to a \textit{babaeng katulong} (\textit{female helper}) than a \textit{lalaking katulong} (\textit{male helper}). Meanwhile, the grammatical markers \textit{ni} and \textit{ang} had positive bias attribution scores, indicating that these induced the model to act with less bias. These results imply that perhaps when the topic concerns power dynamics and relations—as signaled by \textit{pinagsasabihan} (\textit{reprimand}) and \textit{katulong} (\textit{helper})—\texttt{roberta-tagalog-base} might have sexist biases that prompt it to characterize subordinate roles (e.g., \textit{helper}) as female.

Table \ref{tab:bias_attrib2}, on the other hand, presents the bias attribution of the shared tokens in the second challenge prompt entry in Table \ref{table:sample_sentences} as applied to \texttt{sea-lion-3b}. The token with the most negative bias attribution score is \textit{nakikipagtalik} (\textit{fornication}). This score suggests that the word’s presence contributed the most to the model choosing the version of the sentence that associates gay people with promiscuity rather than the version with the straight male subject. These sample analyses illustrate how interpretability analysis using the bias attribution score can improve understanding of how multilingual models operate with bias—especially those handling Filipino texts.

\subsection{Characterizing Bias-Contributing Tokens}
Table \ref{tab:semantic} lists the semantic fields with the ten biggest proportions of bias-contributing words for the models we examined. There are three proportion metrics for each semantic field: [a] the proportion of words in the category with a negative $b(u)$ that increase PLM bias (↑ bias), [b] the 
proportion of words in the category with a positive $b(u)$ that detracted from PLM bias (↓ bias), and [c] the proportion of tokens that got $b(u) = 0$ and had no effect on PLM bias (\(\circ\) bias). The categories in Table \ref{tab:semantic} reveal that there are several semantic fields which provoke biased behavior across all or most of the four PLMs.

One category is that of relationships, which consist of tokens that induce bias 50\% to 60\% of the time on all four models. Words from Filipino CrowS-Pairs that belong to this category are 
\textit{kaibigan} (\textit{friend}), \textit{kasintahan} (\textit{lover}), and \textit{kakilala} (\textit{acquaintance}), hinting that models learned about gender- and sexuality-based biases related to Filipino cultural relationships from their 
pretraining data. The second prompt pair entry in Table \ref{table:sample_sentences} is an example of a sentence in which a 
relational word \textit{nakikipagtalik} (\textit{fornicating}) prompted biased behavior.

Words referring to people (such as \textit{doktor} or \textit{doctor}, \textit{sundalo} or \textit{soldier}, and \textit{katulong} or \textit{helper}) and objects (namely \textit{singsing} or \textit{ring}, \textit{pinggan} or \textit{plate}, and \textit{kandila} or \textit{candle}) also seem to cause models to act with bias. Their effects are particularly potent in \texttt{roberta-tagalog-base} and \texttt{sea-lion-3b}, where they induce bias 45\% to 80\% of the time. The example in Table \ref{tab:bias_attrib1} demonstrates this effect, in which the word \textit{katulong} (\textit{helper}) was among the tokens that prompted \texttt{roberta-tagalog-base} to determine that \textit{Mr. Reyes always reprimands his \textbf{female} helper.} (translated from Filipino) is a more plausible linguistic construction than \textit{Mr. Reyes always reprimands his \textbf{male} helper.}

The concrete and entity-based natures of these bias-contributing categories for Filipino models mark a stark departure from the more abstract and action-based categories that induce bias in 
English models. Whereas \citet{gamboa-2024-interpretability} found that criminal, intimate, and prosocial actions (e.g., \textit{molest}, \textit{raped}, \textit{kiss}, \textit{caring}, and \textit{nurturing}) drive English models to behave with bias, we find that for Filipino models, tangible nouns (e.g., objects and people) have a larger impact on model bias. This insight points to important sociolinguistic differences in how multilingual models handle sociodemographic-related texts written in different languages. 

\begin{table*}[!htbp]
  \centering
  \begin{tabular}{lcccc}
    \hline
    \textbf{Word} & \textbf{Translation} & \textbf{$b(u)$} & \textbf{Direction} & \textbf{Tag(s)}\\
    \hline
    Laging & frequently & $-0.0059$ & more bias & Frequency \\
    pinagsasabihan & reprimand & $-0.0065$ & more bias & Speech acts \\
    ni & \textit{marker} & $0.0064$ & less bias & \textit{stop word} \\
    Ginoong & Mister & $-0.0003$ & more bias & People: Male \\
    Reyes & Reyes & $1.97 \times 10^{-5}$ & less bias & Personal names \\
    ang & \textit{marker} & $0.0078$ & less bias & \textit{stop word} \\
    niyang & his & $0.0012$ & less bias & Pronoun \\
    katulong & helper & $-0.0032$ & more bias & People \\
    \hline
  \end{tabular}
  \caption{Bias attribution scores explaining how the tokens contributed to \texttt{roberta-tagalog-base} choosing the more stereotypical version of this statement over the less biased iteration. }
  \label{tab:bias_attrib1}
\end{table*}

\begin{table*}[!htbp]
  \centering
  \begin{tabular}{lcccc}
    \hline
    \textbf{Word} & \textbf{Translation} & \textbf{$b(u)$} & \textbf{Direction} & \textbf{Tag(s)}\\
    \hline
    Tipikal & typical & $1.11 \times 10^{-8}$ & less bias & Comparing: usual/unusual \\
    siyang & he & $-2.15 \times 10^{-8}$ & more bias & Pronoun \\
    nakikipagtalik & fornicating & $-0.0315$ & more bias & Relationship \\
    kung & if & $0.0019$ & less bias & \textit{stop word} \\
    kani-kanino & anyone & $-0.013$ & more bias & Pronouns \\
    at & and & $-0.040$ & more bias & \textit{stop word} \\
    naghahanap & finding & $-0.0011$ & more bias & Wanting, planning, choosing \\
    lagi & frequently & $-0.0003$ & more bias & Frequency \\
    ng & \textit{marker} & $-0.0217$ & more bias & \textit{stop word} \\
    bagong & new & $0.0415$ & less bias & Time: old, new, and young \\
    jojowain & partner & $-0.0129$ & more bias & Relationship \\
    \hline
  \end{tabular}
  \caption{Bias attribution scores explaining how the tokens contributed to \texttt{sea-lion-3b} choosing the more stereotypical version of this statement over the less biased iteration.}
  \label{tab:bias_attrib2}
\end{table*}

\begin{table*}[!htbp]
\scriptsize
  \centering
  \begin{tabular}{l c c c l c c c}
    \hline 
    
    \hline 
    \multicolumn{4}{c}{\texttt{gpt2}} & 
    \multicolumn{4}{c}{\texttt{roberta-tagalog-base}} \\
    \textbf{Tag} & \textbf{↑ bias} & \textbf{\(\circ\) bias} & \textbf{↓ bias} & 
    \textbf{Tag} & \textbf{↑ bias} & \textbf{\(\circ\) bias} & \textbf{↓ bias} \\
    \hline
    Clothes and personal belongings	&	72.73	&	18.18	&	9.09	&	
    \textbf{People: female}	&	80.00	&	0.00	&	20.00	\\
    \textbf{Relationship: General}	&	54.55	&	9.09	&	36.36	&	Frequency	&	73.68	&	0.00	&	26.32	\\
    \textbf{Objects generally}	&	52.94	&	17.65	&	29.41	&	Knowledge	&	72.73	&	0.00	&	27.27	\\
    Living creatures generally	&	52.00	&	16.00	&	32.00	&	Languauge, speech, and grammar	&	70.00	&	0.00	&	30.00	\\
    Comparing: similar/different	&	50.00	&	16.67	&	33.33	&	Weapons	&	66.67	&	0.00	&	33.33	\\
    Grammatical bin	&	46.67	&	43.33	&	10.00	&	\textbf{Relationship: Intimate/sexual}	&	65.00	&	0.00	&	35.00	\\
    Helping/hindering	&	45.00	&	30.00	&	25.00	&	People	&	64.62	&	0.00	&	35.38	\\
    Being	&	44.44	&	55.56	&	0.00	&	\textbf{Relationship: General}	&	61.54	&	0.00	&	38.46	\\
    Moving, coming, going	&	44.44	&	44.44	&	11.11	&	General appearance	&	60.00	&	0.00	&	40.00	\\
    \textbf{People}	&	44.26	&	16.39	&	39.34	&	\textbf{Objects generally}	&	60.00	&	0.00	&	40.00	\\
    \\
    \hline

    \hline

    \multicolumn{4}{c}{\texttt{sea-lion-3b}} & 
    \multicolumn{4}{c}{\texttt{SeaLLMs-v3-7B-Chat}} \\
    \textbf{Tag} & \textbf{↑ bias} & \textbf{\(\circ\) bias} & \textbf{↓ bias} & 
    \textbf{Tag} & \textbf{↑ bias} & \textbf{\(\circ\) bias} & \textbf{↓ bias} \\
    \hline
    \textbf{Relationship: General}	&	58.33	&	16.77	&	25.00	&	Comparing: Similar/different	&	58.33	&	25.00	&	16.77	\\
    \textbf{People: Female}	&	57.14	&	28.57	&	14.29	&	\textbf{Relationship: General}	&	54.55	&	18.18	&	27.27	\\
    Work and employment	&	57.14	&	33.33	&	9.52	&	Time: Beginning and ending	&	52.63	&	36.84	&	10.53	\\
    Investigate, test, search	&	53.33	&	20.00	&	26.67	&	\textbf{People}	&	51.52	&	24.24	&	24.24	\\
    Business: Selling	&	50.00	&	25.00	&	25.00	&	Business: Selling	&	50.00	&	30.00	&	20.00 \\
    Seem	&	50.00	&	30.00	&	20.00	&	Living creatures generally	&	47.62	&	28.57	&	23.81	\\
    Helping/hindering	&	47.37	&	36.84	&	15.79	&	Speech: Communicative	&	46.15	&	38.46	&	15.38	\\
    \textbf{Objects generally}	&	47.06	&	29.41	&	23.53	&	Kin	&	42.86	&	30.95	&	26.19	\\
    Architecture	&	46.67	&	26.67	&	26.66	&	Calm, violent, angry	&	41.67	&	25.00	&	33.33	\\
    Clothes and personal belongings	&	46.15	&	23.08	&	30.77	&	Time: old, new, and young	&	41.18	&	23.53	&	35.29	\\
    \hline

  \end{tabular}
  \caption{Semantic categories with largest proportions of bias-contributing tokens for the 4 PLMs we examined. ↑ bias: token proportion with $b(u) < 0$ that induced biased behavior. \(\circ\) bias: token proportion with $b(u) = 0$ that did not affect model bias. ↓ bias: token proportion with $b(u) > 0$ that inhibited biased behavior. Categories that induced bias across multiple models are in bold.}
  \label{tab:semantic}
\end{table*}

\section{Conclusion}
\label{sec:conclusion}
In this paper, we extended an existing bias interpretability method for use on models handling 
agglutinative languages like Filipino. Our adjustment of the bias attribution score calculation 
approach emanated from a careful understanding of the morphological differences between 
Filipino, an agglutinative language, and English, an analytic language. We then applied our 
revised method on four models evaluated for bias using Filipino CrowS-Pairs and demonstrated 
the technique’s effectiveness in making transparent how some tokens cause black-box models to 
make biased decisions. Finally, we performed an aggregate analysis of Filipino bias-contributing 
tokens, focusing specifically on the semantic categories they belonged to. Our results show that 
contrary to the abstract and action-heavy nature of bias-contributing tokens in English 
benchmarks and models, Filipino models are induced to act biasedly by words referring to 
concrete entities (i.e., objects and persons). We hope these findings can contribute to current 
efforts investigating bias mechanisms in language models and working to reduce their toxic and 
harmful effects (e.g., \citealp{liu2024devil,ermis-etal-2024-one,gupta2025findingparetotradeoffsfair}.

\section*{Limitations}
Despite broadening the range of languages the bias attribution score method has been applied to, 
our study is still limited to the Filipino language only. While our adjustment of the 
aforementioned approach might be beneficial towards similar agglutinative languages, there 
might still be specificities in other languages and language families that need to be considered 
when the method is applied towards them. These factors therefore need to be considered in future 
work extending the method to other languages.

Our use of the \texttt{googletrans} package to machine translate Filipino tokens before tagging the 
English adaptations using \texttt{pymusas} might have also led to inaccuracies. However, this 
methodological decision was undertaken due to the unavailability of a Filipino semantic tagger 
tool. The development of such a tool in the future may thus be followed by a replication of this 
study for better cultural and linguistic accuracy.

Lastly, the small selection of models we tested our method on is also a limitation of our work. 
We evaluate only four models and do not look into bigger models such as the 7- and 8-billion-parameter versions of SEALION. Moreover, we only include open-source models and are unable to account for proprietary PLMs.

\section*{Acknowledgments}
Lance Gamboa would like to thank the Philippine government's Department of Science and Technology for funding his doctorate studies.

\bibliography{custom}

\appendix

\section{Linguistic Glosses for Sample Prompts}
\label{app:glosses}

\textbf{Dimension:} gender

\noindent\textbf{Bias Profile:} sentence with more bias

\noindent\textbf{Filipino prompt with linguistic gloss: }

\begingl
\gla Laging pinagsasabihan ni Ginoong Reyes ang \textbf{babae} niyang katulong. //
\glb always being.reprimanded by Mr. Reyes the \textbf{female} his helper //
\endgl

\noindent\textbf{English translation:}

\noindent Mr. Reyes always reprimands his \textbf{female} helper. 

\vspace{1cm}

\noindent\textbf{Dimension:} gender

\noindent\textbf{Bias Profile:} sentence with less bias

\noindent\textbf{Filipino prompt with linguistic gloss: }

\begingl
\gla Laging pinagsasabihan ni Ginoong Reyes ang \textbf{lalaki} niyang katulong. //
\glb always being.reprimanded by Mr. Reyes the \textbf{male} his helper //
\endgl

\noindent\textbf{English translation:}

\noindent Mr. Reyes always reprimands his \textbf{male} helper. 

\vspace{1cm}

\noindent\textbf{Dimension:} sexual orientation

\noindent\textbf{Bias Profile:} sentence with more bias

\noindent\textbf{Filipino prompt with linguistic gloss: }

\begingl
\gla Tipikal siyang \textbf{bakla}, nakikipagtalik kung kani-kanino at naghahanap lagi ng bagong jojowain. //
\glb typical 3SG.\textsc{link} \textbf{gay} engaging.in.sex with anyone and seeking always \textsc{gen} new partner.to.date //
\endgl

\noindent\textbf{English translation:}

\noindent He’s a typical \textbf{gay} man, always fornicating with everyone and always looking for a new boyfriend. 

\vspace{2.8cm}

\noindent\textbf{Dimension:} sexual orientation

\noindent\textbf{Bias Profile:} sentence with more bias

\noindent\textbf{Filipino prompt with linguistic gloss: }

\begingl
\gla Tipikal siyang \textbf{lalaki}, nakikipagtalik kung kani-kanino at naghahanap lagi ng bagong jojowain. //
\glb typical 3SG.\textsc{link} \textbf{man} engaging.in.sex with anyone and seeking always \textsc{gen} new partner.to.date //
\endgl

\noindent\textbf{English translation:}

\noindent He’s a typical \textbf{straight} man, always fornicating with everyone and always looking for a new boyfriend. 

\end{document}